\documentclass[10pt,twocolumn,letterpaper]{article}

\usepackage{iccv}
\usepackage{times}
\usepackage{epsfig}
\usepackage{graphicx}
\usepackage{amsmath}
\usepackage{amssymb}
\usepackage{multirow}
\usepackage{array} 
\usepackage{mwe} 
\usepackage{soul}

\usepackage[breaklinks=true,bookmarks=false]{hyperref}

\iccvfinalcopy 

\def\iccvPaperID{****} 

\ificcvfinal\fi

\begin{document}

\title{AI-ready Snow Radar Echogram Dataset (SRED) for climate change monitoring}




\author{Oluwanisola Ibikunle, Hara Talasila, Jilu Li, John Paden\\
Center for Remote Sensing and Integrated Systems (CReSIS)\\
University of Kansas\\
{\tt\small ibikunle.oluwanisola@ku.edu}
\and
Debvrat Varshney\\
Department of Information Systems\\
University of Maryland Baltimore County\\
{\tt\small dvarshney@umbc.edu}
\and
Maryam Rahnemoonfar\\
Department of Computer Science and Engineering  \\
Lehigh University\\
{\tt\small maryam@lehigh.edu}
\thanks{Corresponding author: maryam@lehigh.edu}
}


\maketitle
\ificcvfinal\thispagestyle{empty}\fi

\begin{abstract}
  Tracking internal layers in radar echograms with high accuracy is essential for understanding ice sheet dynamics and quantifying the impact of accelerated ice discharge in Greenland and other polar regions due to contemporary global climate warming. Deep learning algorithms have become the leading approach for automating this task, but the absence of a standardized and well-annotated echogram dataset has hindered the ability to test and compare algorithms reliably, limiting the advancement of state-of-the-art methods for the radar echogram layer tracking problem. This study introduces the first comprehensive ``deep learning ready'' radar echogram dataset derived from Snow Radar airborne data collected during the National Aeronautics and Space Administration Operation Ice Bridge (OIB) mission in 2012. The dataset contains 13,717 labeled and  57,815 weakly-labeled echograms covering diverse snow zones (dry, ablation, wet) with varying along-track resolutions. To demonstrate its utility, we evaluated the performance of five deep learning models on the dataset. 
  Our results show that while current computer vision segmentation algorithms can identify and track snow layer pixels in echogram images, advanced end-to-end models are needed to directly extract snow depth and annual accumulation from echograms, reducing or eliminating post-processing. The dataset and accompanying benchmarking framework provide a valuable resource for advancing radar echogram layer tracking and snow accumulation estimation, advancing our understanding of polar ice sheets response to climate warming. 
\end{abstract}

\section{Introduction}

The continuous monitoring of the Earth's cryosphere, especially the polar ice sheets,  is crucial for understanding its response to global climate warming. Among the ice sheets, the Greenland Ice Sheet (GrIS) is known to be the largest contributor to sea-level rise (SLR) and has become a topic of research since the 1960s \cite{ragle1962south}. Research efforts have identified surface mass balance (SMB) as the main driver of increased GrIS mass loss, making it essential to closely monitor SMB-related processes that are difficult to disentangle \cite{Rignot2019}.

The annual net mass balance of the ice sheet fluctuates in response to SMB-related processes such as snowfall, rainfall, run-off of rainwater and meltwater, sublimation, and evaporation, which are all influenced by annual climatic conditions. As such, the variations in annual climatic conditions are captured in the yearly accumulation.  Therefore, accurately estimating the annual accumulation rate is highly desirable to improve existing climate models used to project future SLR.

\begin{figure}[!t]
\centering
\includegraphics[ width=.65\linewidth, height=.45\textheight]{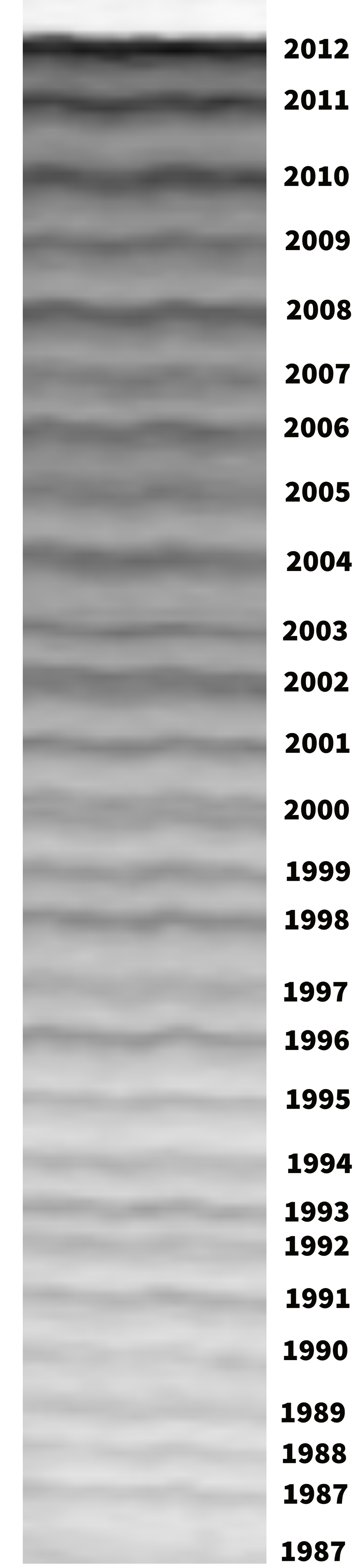}

\caption{Enhanced Snow Radar echogram collected in 2012, illustrating presumed annual accumulation stratigraphy} 
\label{fig:snowradar}
\end{figure}

Snow accumulation rate is traditionally determined by drilling ice cores and shallow pits across the GrIS.  However, this method has inherent limitations, including restricted access, challenging weather conditions, and high costs.  Furthermore, the rapid spatial variability of snow accumulation in GrIS further complicates the ability of the sparse drilled cores to capture catchment-scale accumulation patterns correctly.

To address this limitation, airborne radars have been developed to remotely collect snow measurements at a much larger spatial scale and relatively cheaper cost. These radar systems offer superior subsurface mapping capabilities compared to other remote sensing methods such as space-borne radars. The Center for Remote Sensing and Integrated Systems (CReSIS) has developed a suite of radar systems to investigate different sections of polar ice sheets such as the ultra-wideband Snow Radar which capture annual snow layers in the firn layers of polar ice sheets. 

The Snow Radar's fine vertical resolution of approximately 4 cm, achieved through its wide 6 GHz bandwidth, allows it to resolve subtle contrasts in snow physical properties associated with layers of different snow deposition ages. These contrasts arise from variations in the snow's dielectric permittivity, density and water content, which influence the reflection of the radar's transmitted electromagnetic signal. As the radar signal propagates through the 
snowpack, these interfaces act as electromagnetic boundaries, returning a portion of the signal to the airborne radar system.

The resulting echograms provide a detailed representation of the snowpack stratigraphy, where each distinct layer corresponds to a specific deposition period, such as an annual accumulation cycle. By accurately identifying and tracking these layers, the annual snow accumulation rate can be inferred, offering valuable insights into spatial and temporal patterns of snow deposition. This information is critical for understanding changes in the mass balance of the surveyed polar ice sheet and their contribution to global sea level rise.

Fig. \ref{fig:snowradar} shows a short section of an example radar echogram from the center of the Greenland ice sheet where the accumulation stratigraphy is preserved and the layers correspond roughly to annual accumulation \cite{Koenig2016, Medley2013}. This shows decades of snow accumulation patterns visibly underneath the ice surface. 

To harness the climatic information captured in radar echogram images, robust layer-tracking algorithms are required to automatically detect transitions and trace snow layers. Deep Learning (DL) algorithms, particularly those in the domain of Computer Vision (CV), are well-suited to this task due to their demonstrated success in optical images analysis \cite{deeplabv3+, alexnet,ronneberger2015u} and other remotely-sensed data such as satellite SAR images \cite{cheng2020remote, yuan2021review}. Unlike classical signal processing methods, which rely on hand-crafted feature modeling and require frequent retuning to adapt to variations in snow accumulation pattern across datasets, deep learning models have the potential to generalize effectively. When trained on sufficiently large and diverse datasets, these models can achieve high accuracy in tracking snow layers in echograms from different polar regions and polar ice sheets.

As a result, several deep learning algorithms have been developed to address variants of the layer tracking problem leveraging different training and testing sets. However, the absence of a unified, well-defined training, test, and evaluation set poses significant challenges for meaningful performance comparisons across models. This lack of standardization hinders the identification of best practices required to achieve state-of-the-art results in snow layer detection and tracking. These challenges are further compounded by inconsistencies in evaluation metrics, with some metrics poorly aligned with the ultimate goal of accurately estimating annual snow accumulation.

In this work, we present the Snow Radar Deep Learning Dataset, which provides a standardized resource for training, testing, and benchmarking algorithms.

To evaluate deep learning models, we cast the radar echogram layer tracking problem as a supervised binary semantic segmentation and a deep-tiered multi-class image segmentation task. In this paradigm, we train and evaluate five well-established segmentation algorithms on the test set that contains from echograms from different snow zones and image quality.  This analysis provides insights into the strengths and limitations of each algorithm. Importantly, we assess model performance using evaluation metrics directly linked to physical snow accumulation rates, ensuring alignment with the underlying scientific objectives. 

\section{Related work}

In over three decades of collecting remote radar measurements over polar ice sheets, a large amount of data has been collected to produce a large collection of echograms from different snow zones from Greenland, Antarctica, and Alaskan glaciers. However, the pace of extracting historical climatic information in the echograms has been largely impeded by the lack of efficient and fully-automated snow/ice layer tracking algorithms which are needed to replace the prevalent yet ineffective manual and semi-automated annotation methods \cite{fahnestock2001internal,koenig-accumulation-rate,mitchell2013semi}. As a result, a number of scholarly works exploiting different characteristics of snow surface and subsurface targets have been developed. These approaches can be grouped broadly into three streams: Semi-automated vs fully-automated, surface-bedrock vs internal layer tracking, and machine/deep learning-based vs Non-ML algorithms. 

Semi-automated methods require some form of human input for tracking while fully-automated algorithms are designed to be end-to-end. Surface-bedrock models aim to accurately track only the surface (air-snow boundary) and the bottom (bedrock) layers while layer trackers aim to track laterally persistent layers that could appear at  any depth other than just the surface and bedrock. The surface-bedrock vs internal layer tracking dichotomy is often a result of different radar systems with different vertical resolutions where for some radars, only the surface and the bottom can be clearly seen in their echograms whereas, for others, all the firn layers can be seen and need to be tracked. Finally, non-ML models deploy methods such as statistical models, level-set, probabilistic graphical methods, etc. while machine/deep learning methods develop artificial neural networks to track the desired layers. 

In \cite{gifford2010automated}, the authors combined active contour and thresholding edge-based approach to identify contour boundaries while the study in \cite{bruzzone1} developed statistical models for characterizing subsurface backscatter categorized into strong layers, weak layers, low returns, and basal returns.  In another effort, \cite{koenig-accumulation-rate} developed a high-frequency versus low-frequency semi-automated discriminator algorithm to identify peaks in the returned backscatter power. In another effort,\cite{rahnemoonfar2017automatic} applied distance-regularized level-set function to detect the surface and the ice bottom to improve earlier work by \cite{mitchell} and \cite{dePaulOnana}. In another work, \cite{charged-particles} took inspiration from Coulomb's Law of electrostatic force to detect ice surface and bottom boundaries after performing anisotropic diffusion to enhance layer edges.

Another stream of efforts approached the problem as a probabilistic inference problem by developing probabilistic graphical models to detect layer boundaries. \cite{ice-bottom} pioneered the paradigm while  \cite{lee2014estimating} employed Markov-Chain Monte Carlo (MCMC) over the joint distribution of all possible layer targets in the inference problem. \cite{ice-bottom} expanded the scope to include 3D surface and bedrock reconstruction using Markov random fields (MRF) while \cite{berger2018automated} further refined the approach by incorporating domain knowledge as unary and binary loss function terms.

\cite{bruzzone-carrer} introduced machine learning methods by incorporating Support Vector Machine (SVM) with statistical modeling to classify layers, bedrock, and noise. Since then, several machine and deep learning models have been designed and applied to the RELT problem. 

\cite{kamangir2018deep} introduced a hybrid wavelet network for ice boundary detection. \cite{xu2018multi} designed a multi-task network that avoids the use of meta-data using a combination of recurrent neural networks (RNN) and 3D ConvNets to track layers in 3D radar imagery. In \cite{varshney2020deep}, internal layer tracking was introduced as a semantic segmentation task and deep neural networks were applied to the problem whereas \cite{cai2020end} applied ASPP module using a ResNet-101 backbone to detect layer and bedrock interface in echograms created from Antarctica campaigns. Other efforts in \cite{yari2020multi}, \cite{wang2021deep} applied multi-scale transfer learning and tiered-segmentation approaches for internal layer tracking respectively while \cite{rahnemoonfar2020deep} introduced the addition of synthetic data for model training. In \cite{9622135}, physics-driven and GAN methods were used to create simulated data to train a multi-scale network to improve the accuracy of internal layer tracking. \cite{varshney2020deep} combined the layer tracking task and ice thickness estimation using fully convolutional neural networks and extended the approach in \cite{varshney2022learning} by incorporating physics-defined labels. More recently, \cite{ghosh2022transsounder} combined attention modules with the fusion of TransUNet and TransFuse modules to segment the combination of ice layers, bedrock, and noise similar to efforts in \cite{cai2022accurate} with the addition of Multiscale convolution module (MCM) and focal loss for class weight balancing.

The results from various deep learning approaches demonstrate that end-to-end automatic layer tracking is attainable using deep learning algorithms. However, unlike the computer vision domain for optical images, which benefits from a plethora of well-defined datasets tailored to specific problems, there is currently no standardized echogram layer tracking dataset. To address this critical gap, this work introduces a standardized radar echogram dataset designed to facilitate consistent training and evaluation of deep learning models for snow layer tracking.

\section{Snow Radar echogram background}

The Snow Radar \cite{panzer2013ultra,rodriguez2013advanced} is an ultra-wideband system operating in the 2-8 GHz frequency range, designed to image the upper layers of recent and shallow snow with high precision. Utilizing a frequency-modulated continuous wave (FMCW) architecture and a pulse repetition frequency of 2~kHz, the radar achieves a vertical resolution of approximately 4~cm in snow and an along-track resolution of roughly 15~m. These resolutions are sufficient to distinguish surface and near-surface features, including the air-snow interface, interfaces between snow layers of varying ages, and the snow-ice interface. At each interface, variations in electromagnetic properties (e.g., dielectric permittivity) reflect a portion of the transmitted signal back to the radar receiver, creating peaks in the pulse-compressed range profiles. To convert these raw signals into an interpretable echogram image, a series of signal processing techniques are applied. These include hardware pre-summing, digital filtering, coherent noise removal, and deconvolution \cite{Leuschen2011SnowRadar,rodriguez2013advanced}.

\begin{figure*}[htbp]
  \centering
  \begin{minipage}{0.33\textwidth}
  \includegraphics[width=1\columnwidth, height=.35\textheight]{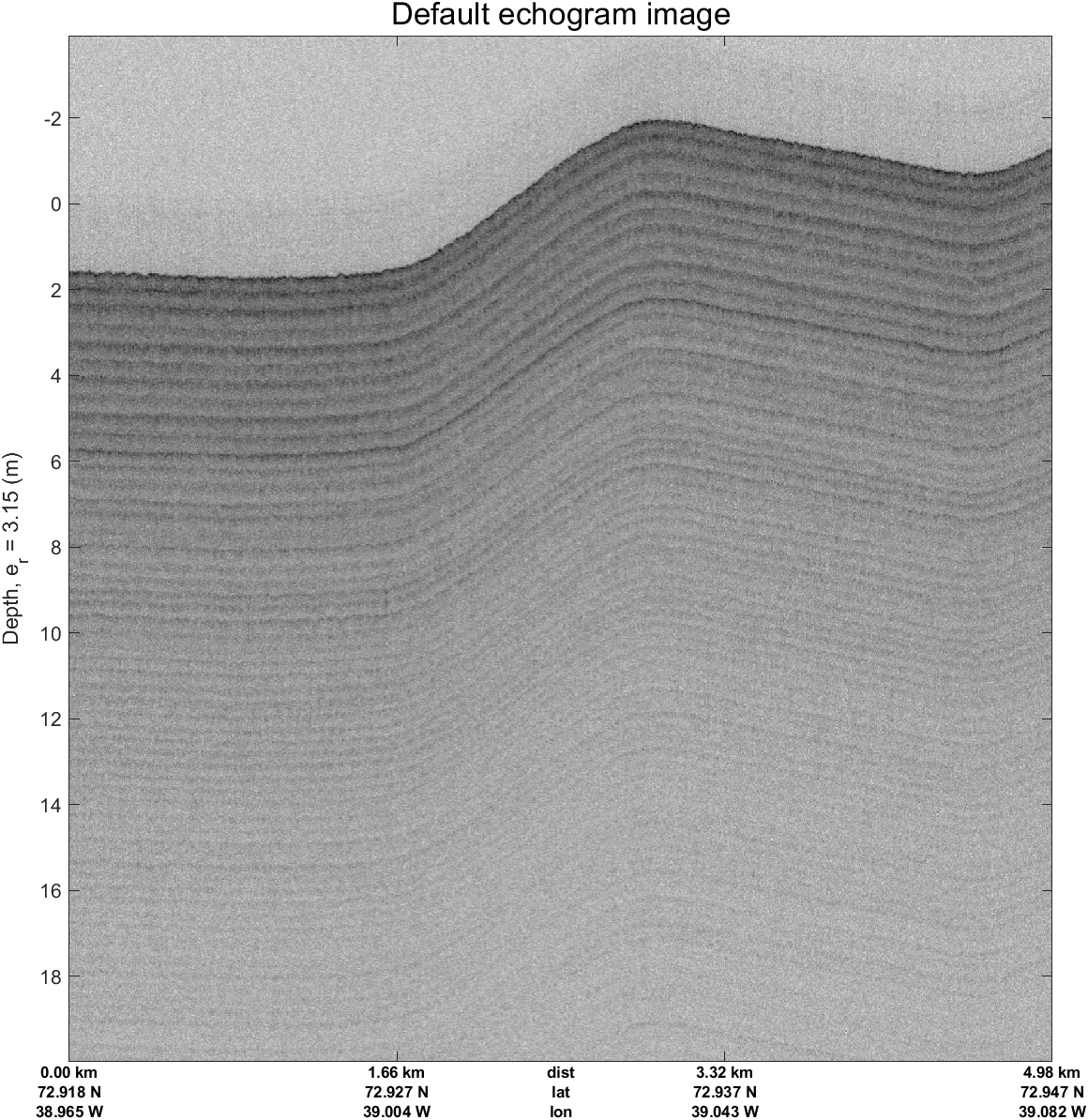}
\end{minipage}%
\begin{minipage}{0.33\textwidth}
  \centering
  \includegraphics[width=1\columnwidth, height=.35\textheight]{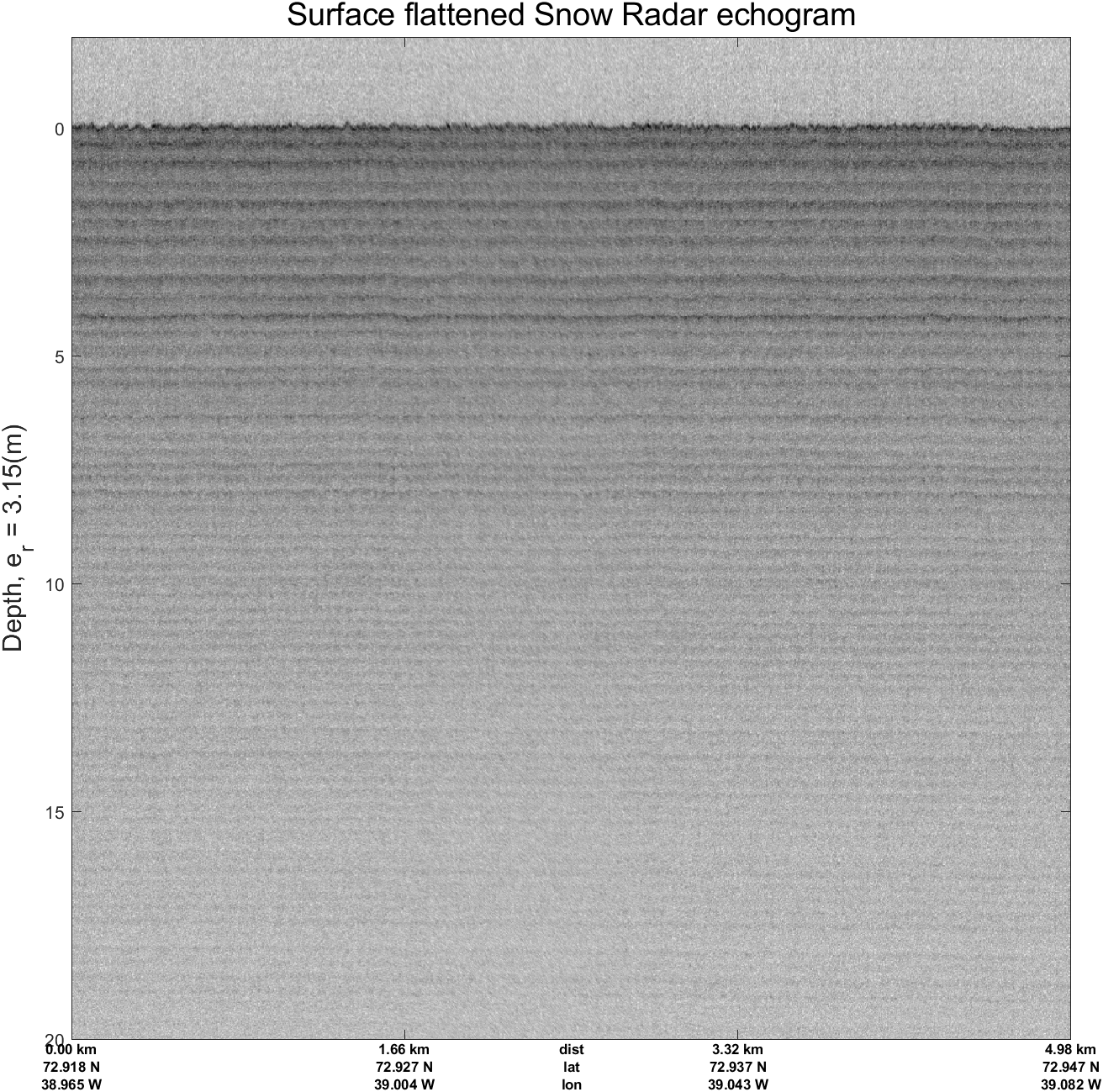}
\end{minipage} 
\begin{minipage}{.33\textwidth}
  \centering
  \includegraphics[width=1\columnwidth, height=.35\textheight]{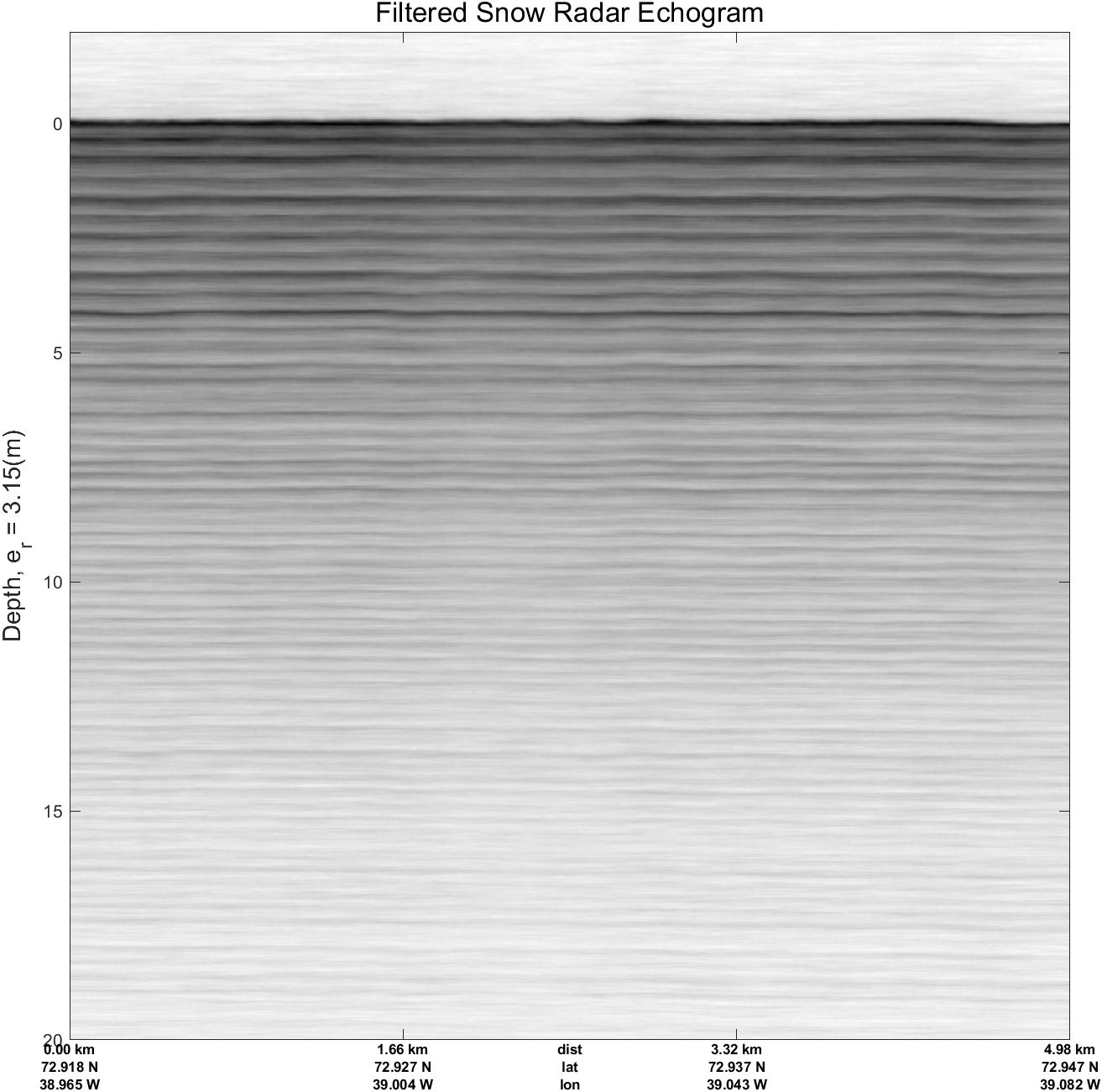}
\end{minipage}

\caption{Plot showing (a) original echogram with topography-elevation effects (b) surface flattened echogram (c) filtered echogram }
\label{fig:default_flat_filtered}
\end{figure*}

Figure \ref{fig:snowradar} shows a surface flattened echogram image. The horizontal axis represents the along-track dimension, corresponding to the direction of aircraft flight, where each column is a `rangeline'. The vertical axis represents the fast-time dimension, where each pixel is a `range bin'. Together, the rangelines (columns) form a 2D echogram image that reveals spatial snow accumulation patterns along the surveyed terrain, while the rows reveal temporal snow accumulation.

The vertical axis is directly related to the depth of each snow layer which can be inferred from the radar's two-way travel time to the layer. The intensity of each pixel in the echogram is determined by the strength of the radar signal scattered back from the resolution cell. 

\subsection{Dataset echogram pre-conditioning}

As can be seen from the echogram image, the inherent characteristics of remotely-sensed data result in visual distinctions from optical images. Phenomena such as signal attenuation due to propagation loss, speckle noise, interference from non-nadir scatterers, and various system imperfections collectively degrade the quality of the echogram. In severe cases, these effects can obscure the deeper snow layers, making them challenging to delineate and track, even for expert human annotators. To mitigate these challenges and enhance the performance of deep learning algorithms in tracking snow layers, we apply a series of critical pre-conditioning steps in the following sequence.

\begin{itemize}
\item {Surface tracking and flattening:} Figure \ref{fig:default_flat_filtered}(a) shows an example echogram prior to surface tracking and flattening. In its unprocessed state, the depth at which the surface (air-snow or air-ice boundary) appears varies depending on the surface relief and topographical changes across the surveyed terrain. This variability makes it challenging to accurately visualize and analyze the isochronous snow layers over long distances. To address this, the dataset echograms are transformed to represent a perfectly flat surface, which is essential for enhancing the visibility of stratigraphic snow layer boundaries and facilitating layer tracking algorithms.

Surface tracking involves identifying the surface bin (range index) in each rangeline of the echogram. This requires an adaptive detection approach because the radar backscatter power and surface return signal exhibit variability across rangelines due to differences in terrain, radar incidence angle, and environmental conditions. We developed an adaptive surface detection algorithm based on the Constant False Alarm Rate (CFAR) principle, augmented with auxiliary data from Digital Elevation Models (DEMs) derived from radar altimetry. The CFAR algorithm dynamically adjusts the detection threshold for each rangeline, accounting for variations in noise and signal strength. By incorporating DEMs, the algorithm further refines surface tracking by leveraging additional topographical context.

After surface tracking, the surface bins are aligned across the echogram using a surface bin alignment algorithm. This step shifts each rangeline's surface bin to a consistent index, effectively flattening the echogram. The result is an image where the surface appears perfectly flat, significantly improving the interpretability of snow layer boundaries. Surface flattening is a critical pre-processing step that facilitates subsequent operations, including along-track filtering and contrast enhancement.

\item {Detrending and 2D filtering:} Due to propagation loss and media attenuation of the transmitted signal as it traverses the ice sheet, returns from deeper snow layers often experience significant signal-to-noise ratio (SNR) degradation. This can make it challenging to detect and track these deeper layers, as their weak signals are buried in noise. To address this, we apply custom detrending methods to enhance the visibility of deeper snow layers by compensating for signal attenuation and remove the power loss as a function of depth trend. This is done by fitting a third-order polynomial to the rangeline's log-power data to remove the trend.

The echogram creation process as detailed \cite{rodriguez2013advanced} already include coherent and incoherent integration to improve data SNR, however, to further improve the tracking performance of the deep learning models, additional mild along-track filtering is applied to the dataset's echogram images. The volumetric backscatter from a snow resolution cell in S-band and C-band frequencies is typically a random process, which introduces stochastic fluctuations in the signal. When this randomness is pronounced, it leads to diffused pixels at the boundaries of snow layers, blurring the layer boundaries and complicating the task of layer detection. These blurred boundaries hinder the ability of deep learning algorithms to distinguish between layers accurately. To mitigate this issue, we apply low-order moving average filters in both dimensions in the linear power domain designed to smooth the data and enhance the contrast between adjacent snow layers, allowing for more precise layer tracking.

\item {Normalization:} Finally, we normalized the pixel values in the echograms to the [0,1] range using a simple linear mapping from backscatter power (in dB) to echogram pixels to ensure quick deep learning model optimization convergence.
\end{itemize}

Figure \ref{fig:default_flat_filtered} shows an example echogram before and after pre-conditioning with improved contrast. 
\section{The Snow Radar echogram deep learning dataset}
The Snow Radar Echogram Deep Learning dataset (SRED) was generated from airborne radar data collected by the Snow Radar over the Greenland Ice Sheet during the NASA's Operation Ice Bridge (OIB) campaign in the spring of 2012. As is common with remotely-sensed data collection campaigns, a large number of echograms have been produced from the data but they lacked labels. To provide the research community with a standardized dataset for deep learning model training, the SRED dataset was curated, consisting of a manually labeled set and a weakly-labeled echograms, the latter created via model inference. The labeled subset of the dataset includes  11,302 echograms designated for training, 1,323 for validation, and and 1292 for testing. These echograms cover data collected from 11 flight lines spanning about 30,000~km along the surveyed area of the Greenland Ice Sheet. This dataset captures a variety of snow zones allowing the training of models that can generalize across diverse snow conditions. The SRED dataset is publicly available for research purposes and can be accessed at \href{https://gitlab.com/openpolarradar/opr/-/wikis/Machine-Learning-Guide}{https://gitlab.com/openpolarradar/opr/-/wikis/Machine-Learning-Guide}.

The number of snow layers in an echogram is not predefined and varies based on the topography and climatic conditions at the data collection site. These conditions influence snow accumulation patterns, which are captured in the echograms. Snow zones are commonly categorized into three types: dry, transition, and wet zones. Dry snow zones experience little or no melting, transition zones contain some infiltration of liquid water, and wet zones have significant melting and higher water content. Water acts as a specular reflector of radar signals, and its presence can degrade the quality of echogram images. To construct the training and validation sets for the SRED dataset, data from 9 out of the 11 flight lines were selected, ensuring representation of all identified snow zones and accumulation patterns across Greenland. For testing, one flight line from the dry-snow zone near Greenland’s summit and another spanning all three snow zones were used. This approach allows for an assessment of model performance across diverse snow conditions and provides insights into their generalizability.

Additionally, the flight lines in the SRED dataset are in close proximity to or coincide with existing drilled ice cores and shallow pits, providing an opportunity for further research on how deep learning algorithms can synchronize radar echogram data with climate data. This may offer valuable insights for improving ice core measurements, particularly in terms of ice-age-depth relationships.

\begin{figure}[htbp]

\includegraphics[ width=1.\linewidth, height=.35\textheight]{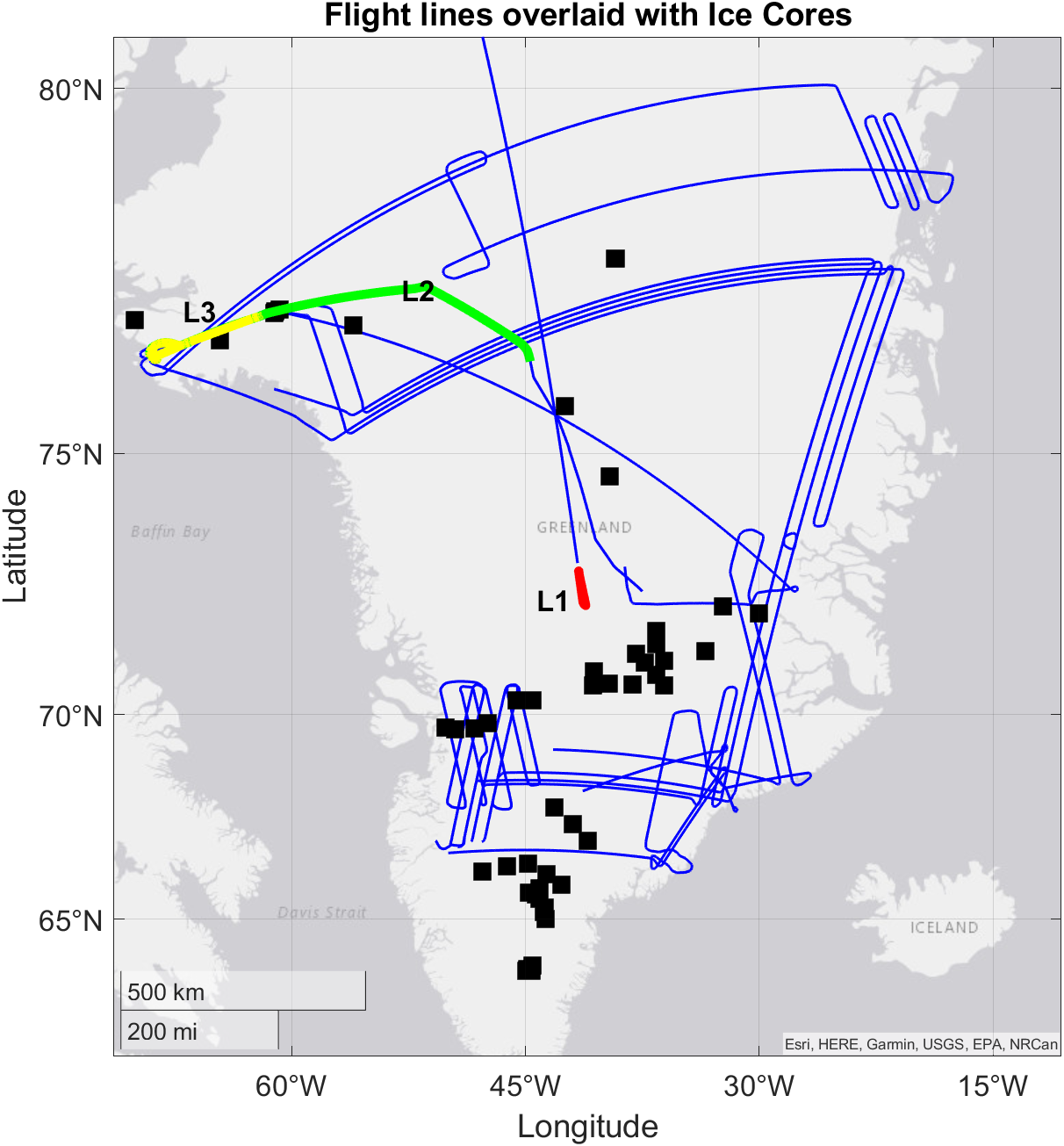}

\caption{Map of Greenland showing the geolocation of flight lines used for dataset echograms and neighboring ice cores. Blue lines indicate the training data flight lines, while red, green, and yellow lines represent the L1, L2, and L3 test sites, respectively while black squares represent ice cores and shallow pits.} 
\label{fig:flight_line_ice_cores}
\end{figure}

Figure \ref{fig:flight_line_ice_cores} is a plot of the dataset flight lines over Greenland and some of the ice cores and shallow pits drilled in Greenland.

\subsection{Dataset echogram resolution}
To investigate how the tracking performance of different deep learning models varies in response to the along-track distance covered by an echogram image, we employed a ``multi-distance-multi-looking'' strategy. This creates echograms images with consistent width and height but spanning different physical along-track distances. The use of different flight line distances to create echogram images for training deep learning models has the advantage of simulating different spatial accumulation patterns seen by the models. This achieves the dual benefit of increasing the diversity in the training data while also increasing the generalizability and robustness of the trained models to ``unseen'' spatial accumulation distribution when used for inference on echograms produced from future surveys.

For this work, we create echogram images that span distances of 2km, 5km, 10km, 20km, and 50km. To ensure uniformity of the dimensions of each training sample, we applied linear interpolation in the along-track axis to create fixed echogram image sizes of 1664 rangebins and 256 rangelines. This dimension uniformity standardizes the echogram inputs to help in the true assessment of the performance of different deep learning models on the dataset. The associated distance of each echogram block is appended to its filename for easy identification.


\begin{table}[ht] 
 \begin{center}   
    \begin{tabular}{l|c c c c c} %
      $\quad$ & \textbf{Train} & $\quad$ & 
      \textbf{Validation} & $\quad$ & \textbf{Test} \\

      \hline
       \textbf{2km} & 6577 & $\quad$ & 689 & $\quad$ & 911\\
       \textbf{5km}  & 2776 & $\quad$ & 359 & $\quad$ & 366\\
       \textbf{10km} & 1299 & $\quad$ & 208 & $\quad$ & 15\\
      
    \end{tabular}
  \end{center}
  \label{table:Training_distrib}
  \caption{Distribution of echograms based on along-track distances in train, validation, and test sets }
\end{table}

\subsection{Test set split based on snow zones}
The test set is carefully divided into 3 groups: L1, L2, and L3 which partly represent different snow accumulation patterns that exist in most polar ice sheets. The L1 test set is from the dry snow zone with well-preserved annual accumulation stratigraphy. L3, on the other hand, is close to the coast and contains echograms from the wet snow zone where layer stratigraphy might have been eroded due to melting while the L2 test echograms capture the transition between both zones. The quality of the echograms produced (i.e visual appearance and ability to see each snow layer clearly) varies appreciably depending on whether the data is from the dry or wet snow zone. The L1 section of the test data has the best image quality compared to the echograms in L2 and L3 sections. This is because the radar backscatter from L2 and L3 zones is diffused and scattered away from the radar receiver due to the presence of melted snow water which appears as a smooth surface that reflects the electromagnetic wave away from the radar. This split of the test data is done to further investigate how deep learning models perform depending on input echograms of varying echogram quality from different snow zones.

\begin{table}[ht] 
 \begin{center}   
    \begin{tabular}{l|c c c c c} %
      $\quad$ & \textbf{L1} & $\quad$ & 
      \textbf{L2} & $\quad$ & \textbf{L3} \\

      \hline
       \textbf{2km} & 81 & $\quad$ & 752 & $\quad$ & 78\\
       \textbf{5km}  & 32 & $\quad$ & 297 & $\quad$ & 38\\
       \textbf{10km} & 15 & $\quad$ & - & $\quad$ & -\\      
    \end{tabular}
  \end{center}
  \label{table:Test_distrib}
  \caption{Distribution of echograms in the test set based on snow zones and echogram image quality }
\end{table}

\subsection{Radar Echogram internal layer tracking problem definition}

The Snow Radar echogram images capture the top firn layers of the surveyed location and the goal is to develop custom deep-learning algorithms that can detect and track the snow layers uniquely over long distances so as to estimate the annual accumulation rate. 

Formally, given an input 2D grayscale echogram image \textbf{E} $\in$  $\{ \mathbb{R}^{N_t \times N_x}: 0 \leq E(m,n) < 1 \}$ where \textbf{E} represents the two-dimensional spatial distribution of firn layer backscatter in the along-track (or azimuth) direction and in the rangebin (or depth) axis. $N_t$ is the number of range bins (similar to image height) while $N_x$ is the number of range lines (similar to image width) of the echogram image. A deep learning model is to identify which of the 2~D echogram matrix, \textbf{E}, pixels contain a snow layer and track (at most) $N_x$  consecutive columns for each layer in the along-track axis. The output of the algorithm for \textbf{L} unique layers in the test echogram would then be \textbf{O} $\in$  $\{ \mathbb{R}^{N_t \times N_x} \}$. 

In the context of supervised deep learning, this problem can be cast into 2 paradigms:
(I) Binary image segmentation and (II) Deep-tiered multi-layer segmentation. In both cases, an equal-sized corresponding ground truth annotation \textbf{G}  $\in$  $\{ \mathbb{R}^{N_{t} \times N_{x} } \}$ is provided as a supervisory signal to train the models to correctly identify snow accumulation layer pixels.

\begin{figure}[htbp]

\includegraphics[ width=1.\linewidth, height=.35\textheight]{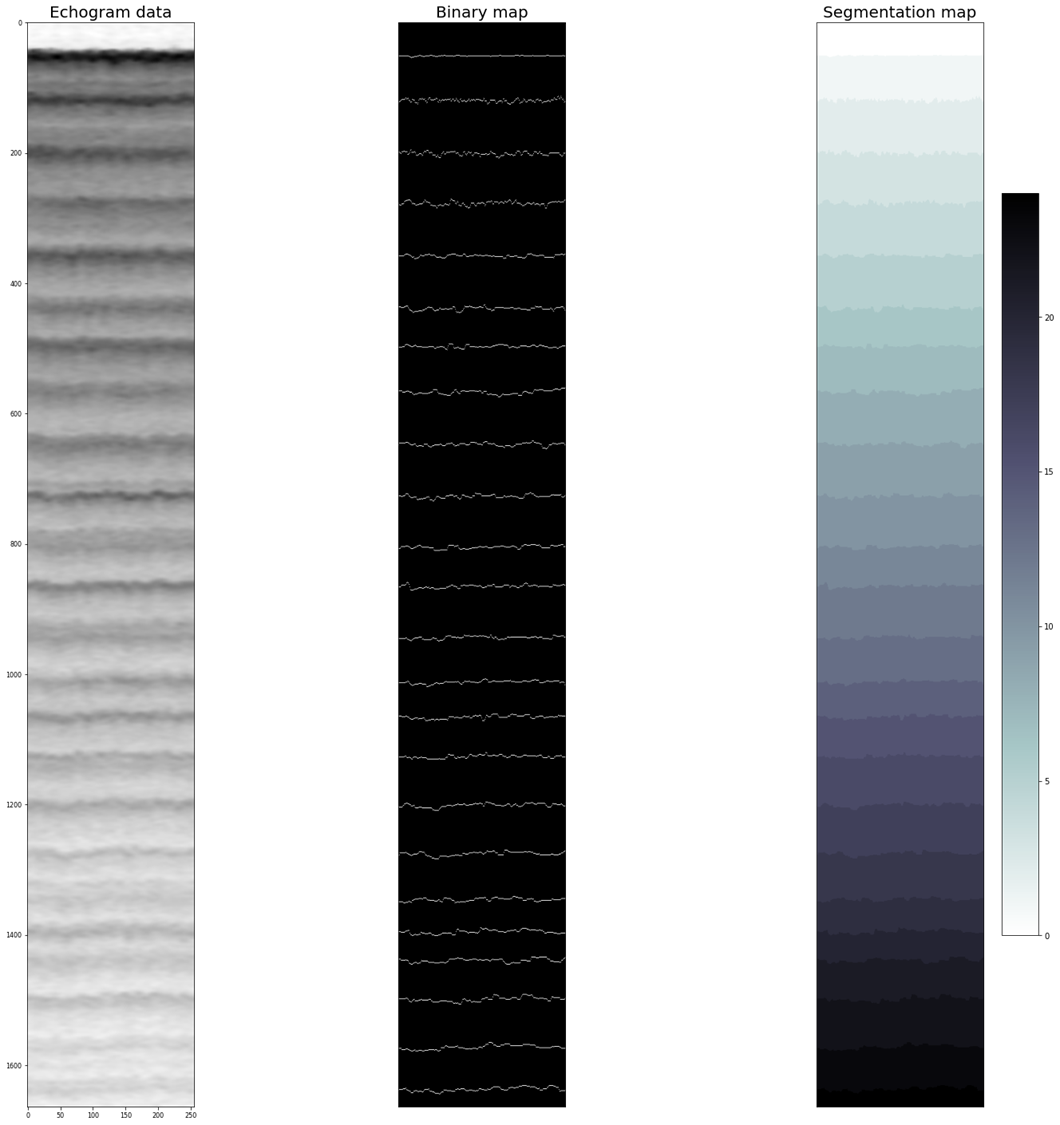}

\caption{(a) Echogram image (b) Binary segmentation map (c) Multi-class segmentation map} 
\label{fig:echogram_binary_multiclass}
\end{figure}

\begin{enumerate}
    \item Binary segmentation: 
    In this paradigm, the model is trained to classify each pixel in the image as either containing a layer (1) or not (0) using the associated ground truth binary matrix annotation of similar dimension with $ G_{b} \in \mathbb{R}^{ N_{t} \times N_{x}}, \; G_{b}(m,n) = \{ 0, 1 \}$. While there can be a variety of inner architecture of the model, the output layer of the binary segmentation neural network generally has a sigmoid activation function that is thresholded to produce binary outputs. However, this binary output  needs to be post-processed to extract each annual layer's 1~D contour in order to estimate the annual accumulation.

    \item Deep-tiered multi-layer segmentation: Given that each layer in the echogram image corresponds to a seasonal snow accumulation (assumed to be annual), the multi-layer segmentation seeks to uniquely identify the pixels attributed to each year's deposition in the input echogram. The most recent accumulation prior to the radar measurement is seen as the first layer while older years are the deeper layers sequentially arranged in chronological order (see Figure \ref{fig:echogram_binary_multiclass}). 
    
    To train a model to uniquely identify pixels associated with each year's accumulation and delineate adjacent accumulation boundaries, the ground truth annotation is coded to uniquely identify each year's stratigraphy. Concretely, ground truth annotation $ G_d \in \{ \mathbb{R}^{N_t \times N_x} : G_d(m,n) \in \{0, 1, 2, \dots, L_{\text{max}} \} \}$ where each pixel is assigned a tiered label L based on its associated year of deposition. $L = 0 $ typifies the signal-in-air portion before the transmit signal interacts with the surface, $L = 1$ represents the first year's accumulation, $L = 2$ represents the second year's accumulation, while $L= L_{max}$ corresponds to the deepest accumulation layer in the input echogram image. 

    This approach is similar to multi-class image segmentation of optical images with the unique difference that in deep-tiered multi-layer segmentation, the accumulation layers are naturally ordered based on the year of the snowfall and a layer only shares horizontal boundaries with adjacent classes (accumulation layers of the year before and after). This approach to the radar echogram layer tracking problem has the advantage of directly estimating each year's annual accumulation rangebins since it uniquely identifies each layer pixel in the along-track axis and delineates its boundaries with adjacent accumulation layers. However, this requires a larger model which comes at the cost of increased parameterization of the model and the inherent need for more training data and compute resources.
    
    \end{enumerate}

In this work, we report the binary image segmentation benchmark because of its relative simplicity, fewer numbers of trainable parameters, and most importantly its ability to detect as many snow layers in an echogram without the need for hard-coding this in the model architecture as in the case of multiclass segmentation models. The goal is to deploy these models in an active learning framework to increase available echograms with correct layer tracking that can be subsequently added to create the desired large dataset.
    \begin{table*}[htbp]
        \centering
        \begin{tabular}{|l|c|c||c|c||c|c|}
            \hline
             \multirow{2}{*} \textbf{Network} & \multicolumn{2}{c||}{\textbf{L1}} & \multicolumn{2}{c||}{\textbf{L2}} & 
             \multicolumn{2}{c|}{\textbf{L3}} \\
             \hline
             & \textbf{ODS} & \textbf{OIS} & \textbf{ODS} & \textbf{OIS} & 
             \textbf{ODS} & \textbf{OIS} \\ 
             \hline 
             UNet  & 0.916 & 0.917 & 0.789 & 0.790 & 0.214 & 0.214\\ 
             \hline             
            AttentionUNet & 0.971 & 0.972 & 0.908 & 0.909 & 0.218 & 0.218\\ 
            
             \hline 
                DeepLabv3+  & 0.959 & 0.960 &	0.880 & 0.881 & 0.159 & 0.159\\ 

             \hline
              FCN & 0.957 & 0.958 & 0.913 & 0.914 & 0.187  & 0.187\\ 
             
             \hline            
             Soft Ensemble & 0.962 & 0.963 & 0.885 & 0.886 & 0.167 & 0.167 \\ 
             
             \hline
            
        \end{tabular}
        \bigskip
        \caption{Optimal Dataset Scale (ODS) and Optimal Image Scale (OIS) for each model for the different snow zones in the test set.}
        \label{tab:OIS_ODS_table}
    \end{table*}

\section{Experiments}

\subsection{Baseline models}
We applied four well-known deep learning models; Fully Convolutional Networks (FCN) \cite{long2015fully}, UNet \cite{ronneberger2015u}, Attention-UNet \cite{oktay2018attention}, DeepLabv3+ \cite{chen2017rethinking} and an additional soft ensemble of the outputs of each of the models. The models were trained in a fully-supervised mode for binary image segmentation task (i.e. to predict if a pixel in the echogram image contains a layer or not).

FCN, UNet, and AttentionUNet share very similar architecture particularly the encoder-decoder structure with repeated spatial pooling  in the encoder followed by corresponding upsampling in the decoder. Importantly, UNet model employs the use of skip connections between the encoder and decoder layers of similar image resolution. AttentionUNet applies a gating mechanism to adaptively weigh the skip connections while FCN architecture avoids the use of skip-connections altogether. DeepLabv3+ combines spatial pyramid pooling with the encoder-decoder architecture effectively reducing the number of short-cut connections needed between the encoder and the decoder.


\subsection{Experimental setup}

For training, we used the standard architectures of the deep learning models with slight modifications for compatibility with radar echograms. The FCN, UNet, and AttentionUNet all have four convolution blocks in the encoder and decoder. Each convolution block consists of sandwiched convolution, batch normalization, dropout, activation, and  downsampling (in the encoder) or upsampling (in the decoder). Our DeepLabv3+ implementation uses dilation rates of 6,12, and 18 in the spatial pyramid pool and the ResNet50 architecture as the backbone with the weights trained from scratch.

All the models were trained with the binary focal loss with $\alpha = 0.25$ and $\beta = 2$ to mitigate the inherent binary class imbalance.

\section{Evaluation and Benchmarking}

The performance of the deep learning models on each section of the test set is evaluated in two stages. In the first stage, the model's direct outputs are evaluated after thresholding and binarizing the network's softmaxed outputs. In the second stage, uniform post-processing to uniquely identify each snow layer contour in the echogram is applied, and the model’s ability to accurately track the snow layers is further evaluated and quantified.

\subsection{ Model binary output evaluation}
\label{sec: binary_outputs }
\begin{figure}[htbp]
    \centering
    \includegraphics[width=0.9\linewidth, height = .40\textheight]{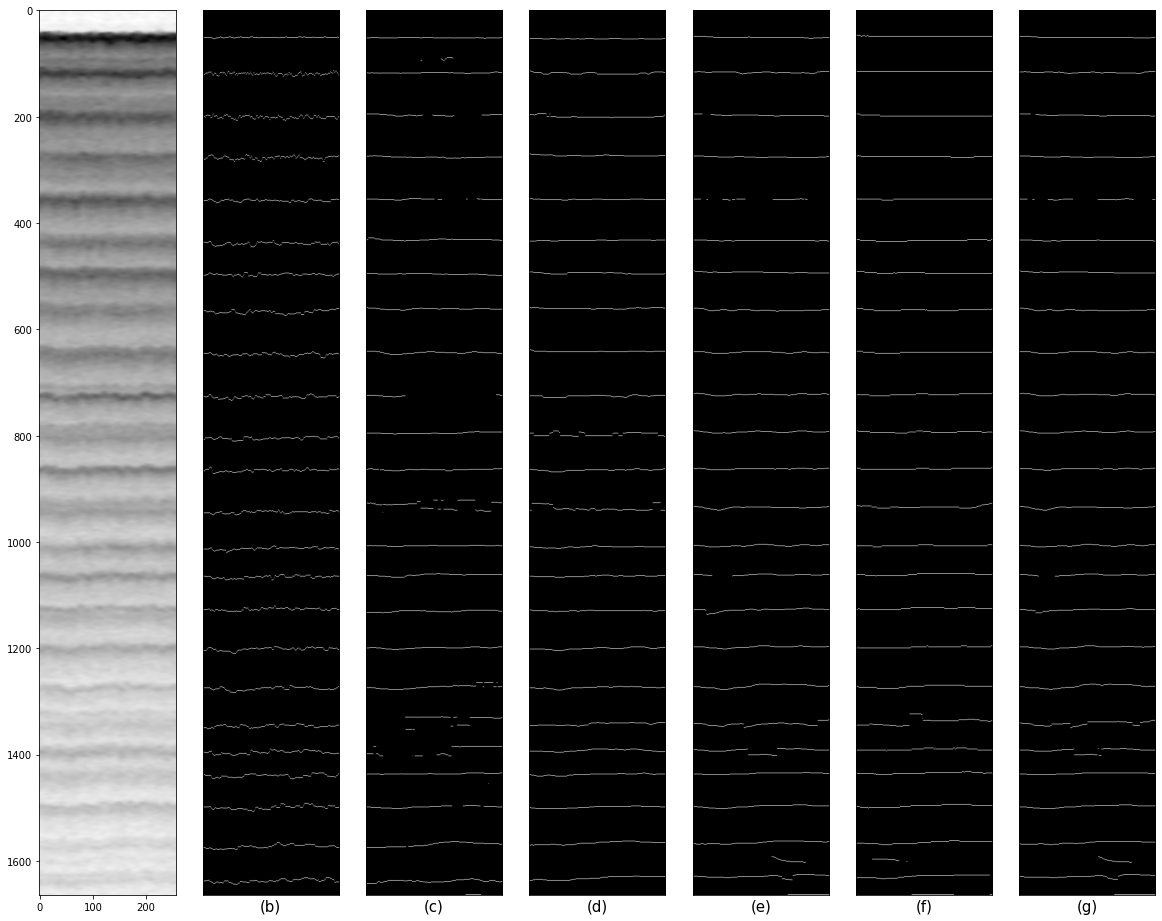}
    \caption{Echogram image and model binary outputs (b) Ground truth annotation (c) UNet (d) AttentionUNet (e) DeepLab (f) FCN  (g) Soft Ensemble }
    \label{fig:echo_binary_qual}
\end{figure}

The immediate outputs of the networks are first thresholded using a simple non-maximum suppressing algorithm to create binary results as shown in Figure \ref{fig:echo_binary_qual}. To evaluate the performance of the models to correctly classify layer pixels in the echogram, we Optimal Dataset Scale (ODS) and Optimal Image Scale (OIS) F-scores as our evaluation metrics. Both metrics are similar but operate over the dataset in two distinct ways. ODS F-score computes the best threshold that yields the optimum F-score over the entire dataset while OIS does this for each individual image and then computes the average of the optimal F-scores for the entire dataset.

\begin{figure}[htbp]
    \centering
    \includegraphics[width=0.9\linewidth, height =.40\textheight]{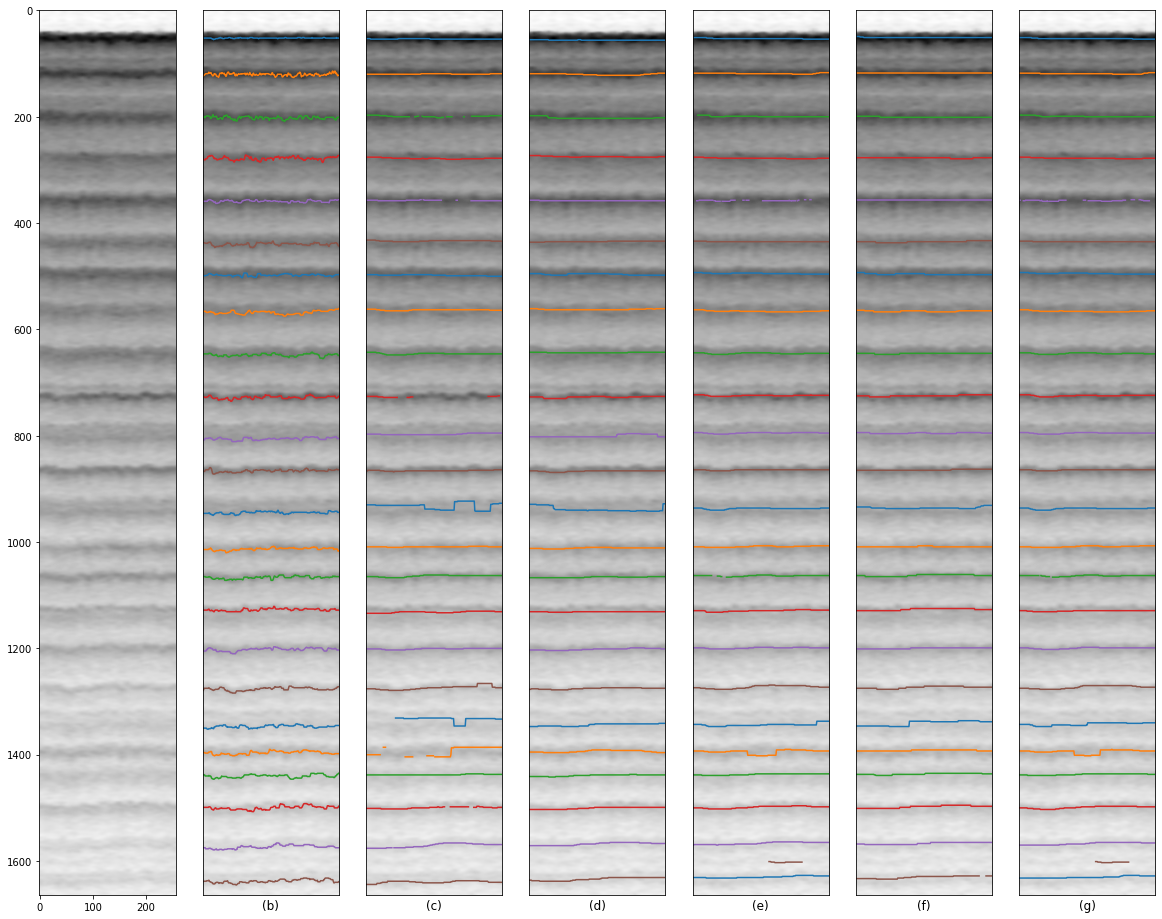}
    \caption{Echogram image and vectorized model outputs (b) Ground truth annotation (c) UNet (d) AttentionUNet (e) DeepLab (f) FCN  (g) Soft Ensemble }
    \label{fig:single_qual}
\end{figure}

\subsection{ Model tracking evaluation}
 The primary objective of the models is to track and estimate annual snow accumulation. To achieve this, the binary raster output of the models (a  matrix of 0's and 1's with dimensions $N_t \times Nx $) must be post-processed to create layer contour matrix ($N_{L} \times N_{x}$) that uniquely identifies  and tracks each annual snow layer in the echogram. This is accomplished through a post-processing routine that extracts the row indices (range bin) of all pixels classified as part of a layer (i.e., a value of 1 in the model's binary output) and clusters them into distinct $N_{L}$ layers. 
 
The clustering process leverages the physical property that snow layers do not intersect due to the continuous accumulation of snow between layers. First, the row indices of all layer-containing pixels are extracted. Then, the algorithm groups these indices into distinct annual layers, ensuring that each $N_L$ layer is uniquely identified. The output is a layer vector ($N_L \times N_x$) for the entire echogram, encoding each snow layer as a single vector. This format closely matches the ground truth layer vectors and enables accumulation rate estimation by calculating the vertical accumulation gap between each tracked snow layer and the next. 

Finally, the performance of each model's layer tracking is evaluated by comparing the model output vector with the corresponding layer vector ground truth.

The metrics reported in Section \ref{sec: binary_outputs } above primarily measures of how well the models classify the echogram pixels into ``layer'' or ``no-layer'' categories. However, these metrics provide limited insight into the models' ability to track layers in along-track and, more importantly, to solve the underlying problem of estimating the annual snow accumulation rate. To address this limitation, we introduce and apply N-pixel accuracy metrics and compute the Mean Absolute Error (MAE) to better evaluate model performance in terms of layer tracking and accumulation rate estimation.

The N-pixel accuracy is a simple metric that calculates the percentage of the predicted layer vectors that exactly match or fall within an N-pixel range of the corresponding ground truth label. It evaluates the proportion of the model's prediction error that is lesser than or equal to a specified \textbf{N} threshold. This metric is particularly relevant because, although the snow layers in echogram images are typically more than one pixel thick, the ground truth labeling process assigns only a single pixel to represent each layer. As such, it gives better evaluation of how well the algorithms track the layers beyond the limitation of the imperfection of the ground truth labels. 

\begin{align}
  \text{N\_pixel\_accuracy} = & \frac{1}{N_{\text{echo}} N_L N_x} 
  \sum_{n=1}^{N_{\text{echo}}} 
  \sum_{l=1}^{N_L} 
  \sum_{j=1}^{N_x} \mathbb{I} 
  \big(|\hat{Y}_{n,l,j} - Y_{n,l,j}| \leq \mathbf{N} \big) \nonumber \\
  & \times |\hat{Y}_{n,l,j} - Y_{n,l,j}|
    \label{eqn:N_px}
\end{align}

\begin{gather}
\text{MAE} = \frac{1}{N_{\text{echo}} N_L N_x} \sum_{n=1}^{N_{\text{echo}}} \sum_{l=1}^{N_L} \sum_{j=1}^{N_x} \left| \hat{Y}_{n,l,j} - Y_{n,l,j} \right|
\label{eqn:MAE_calc}
\end{gather}

In Equation \ref{eqn:N_px}, $\mathbb{I} \{.\}$ is the indicator function that evaluates to one if $\{.\}$ is true and zero otherwise.

$Y_{n,l,j}$  is the labeled ground truth for rangeline (column) $j$ of layer $l$ in echogram $n$ of the entire ${N_{\text{echo}}}$ test echograms. $\hat{Y}_{n,l,j}$ is the corresponding model's prediction. 


Using the accumulation rate-related metrics N-pixel accuracies in  Equation \ref{eqn:N_px}, we report the N-pixel tracking performance of the models in the L1, L2, and L3 test echograms. For this evaluation, we consider thresholds of $N = 2,5,\text{and} \: 10$, providing a flexible measure of the models' layer tracking accuracy under varying tolerances.

    \begin{table*}[htbp]
        \centering
        \begin{tabular}{|l|c|c|c|
c||c|c|c|c||c|c|c|c|}
            \hline
              & \multicolumn{4}{c||}{\textbf{L1}} & \multicolumn{4}{c||}{\textbf{L2}} & 
             \multicolumn{4}{c|}{\textbf{L3}} \\ 
             \hline
             \textbf{Model}
             & \textbf{2px} & \textbf{5px} & \textbf{10px}
             & \textbf{MAE}
             & \textbf{2px} & \textbf{5px} & \textbf{10px}
             & \textbf{MAE}&             
             \textbf{2px} & \textbf{5px} & \textbf{10px} &
             \textbf{MAE}
             \\
             \hline 
             UNet  & 0.141 & 0.719 & 0.973 & 4.222 & 0.108 & 0.53 & 0.831 & 6.872 & 0.001 & 0.077 & 0.225 & 70.988 \\            
             \hline             
            AttentionUNet & 0.269 & 0.806 & 0.986 & 3.466 & 0.110 & 0.719 & 0.939 & 4.970 & 0.001 & 0.050 & 0.174 & 69.266 \\
            
             \hline 
                DeepLabv3+  & 0.124 & 0.761 & 0.987 & 3.956 & 0.175 & 0.637 & 0.916 & 5.118 & 0.001 & 0.022 & 0.175 & 41.661  \\

             \hline
              FCN & 0.095 & 0.788 & 0.979 & 3.982 & 0.325 & 0.805 & 0.963 & 3.742 & 0.001 & 0.084 & 0.278 & 47.084 \\ 
             
             \hline            
             Soft Ensemble & 0.141 & 0.768 & 0.988 & 3.873 & 0.184 & 0.653 & 0.920 & 5.010 & 0.001 & 0.044 & 0.183 & 42.672 \\ 
             
             \hline
            
        \end{tabular}
        \bigskip
        \caption{N-pixel accuracy and MAE for each model for L1, L2, and L3 test sets.}
        \label{tab:N_pixel_acc}
    \end{table*}


We also evaluate the mean absolute error (MAE), as defined in Equation \ref{eqn:MAE_calc}, for each along-track distance in the L1, L2, and L3 test sections. This analysis provides deeper insights into model performance by examining how accurately each model tracks snow layers across varying echogram qualities and spatial extents.

By considering both the N-pixel accuracies and the MAE, we present a comprehensive assessment of each model's robustness and generalizability under diverse test conditions, highlighting their strengths and potential areas for improvement.

To relate the pixel margins and MAE estimates back to the snow layer accumulation tracking estimation physical problem, the layer thickness errors is converted back to meters using the Snow Radar's fast-time sampling rate $\Delta_t = 0.0852~ns$. Assuming a dielectric of $\epsilon_r = 2$ which is in between the fresh fallen snow ($\epsilon_r = \sim 1.5$) and solid ice ($\epsilon_r = \sim 3.15$), the fast-time or row-pixel height is computed:

\begin{equation}
\Delta_r = \frac{c \Delta_t}{2\sqrt{\epsilon_r}} = 0.01 \, \text{m}.
\label{eqn:range_resolution}
\end{equation}

Consequently, the UNet L2 mean absolute error (MAE) of 6.872 corresponds to 62~cm while FCN's MAE of 3.742 equates to 33~cm accumulation tracking error.

\section{Conclusion and Future work}
This paper introduces  the Snow Radar Echogram Deep Learning (SRED) dataset, the first comprehensive, publicly available radar echogram dataset tailored for deep learning applications in snow layer tracking. By addressing the lack of standardized and well-annotated radar echogram datasets, the SRED dataset bridges a critical gap, enabling the evaluation and benchmarking of state-of-the-art deep learning algorithms for internal layer tracking. The dataset spans diverse snow zones and includes multiple along-track resolutions, offering a valuable resource for advancing research in radar echogram analysis.

We evaluated the performance of five prominent deep learning models, highlighting their strengths and limitations under varying snow zone conditions and echogram qualities. Metrics such as N-pixel accuracy and Mean Absolute Error (MAE) were used to quantitatively assess the models' ability to track snow layers and estimate annual accumulation rates. The findings demonstrated that while models such as AttentionUNet and DeepLabv3+ excelled in well-preserved dry snow zones, their performance diminished significantly in wet snow regions due to degraded layer stratigraphy. These results underscore the need for improved architectures and methods capable of handling challenging echogram conditions.

Future work will focus on several key areas to extend the impact of this work. This includes developing advanced end-to-end deep learning architectures to improve layer tracking in low-quality echograms, particularly wet snow zones and also minimize the need for post-processing by directly estimating snow accumulation from echogram images. We intend to use this in the active learning paradigm to include additional flight lines and polar regions, such as Antarctica, to improve model generalizability across diverse geographic and climatic conditions.

\paragraph{Funding Statement}
This study was funded by NSF BIGDATA awards (IIS-1838230, IIS-1838024)

\end{document}